\documentclass[lettersize,journal]{IEEEtran}
\usepackage{amsmath,amssymb,amsfonts}
\usepackage{algorithm}
\usepackage{array}
\usepackage[caption=false,font=normalsize,labelfont=sf,textfont=sf]{subfig}
\usepackage{textcomp}
\usepackage{stfloats}
\usepackage{xurl}
\usepackage{verbatim}
\usepackage{graphicx}
\usepackage{cite}
\usepackage{orcidlink}

\usepackage{amsthm}
\usepackage{url} 
\usepackage{booktabs}

\usepackage{times}
\usepackage{float}
\usepackage[english]{babel}
\usepackage{algorithmicx}%
\usepackage{multirow}%
\usepackage{mathrsfs}%
\usepackage{xcolor}%

\usepackage{array}     
\usepackage{siunitx}   
\usepackage{changepage}

\usepackage{textcomp}%
\usepackage{manyfoot}%
\usepackage{algpseudocode}%
\usepackage{listings}%
\usepackage{afterpage}
\usepackage{pdflscape}
\usepackage{rotating}

\begin{document}
\title{Loss-Guided Model Sharing  and Local Learning Correction  in Decentralized Federated Learning for Crop Disease Classification 
\thanks{This project is funded as part of the MERIAVINO project. MERIAVINO is part of the ERA-NET Cofund ICT-AGRI-FOOD, with funding provided by national sources [ANR France, UEFISCDI Romania, GSRI Greece] and co-funding by the European Union’s Horizon 2020 research and innovation program, Grant Agreement number 862665.}}

\author{
Denis MAMBA KABALA\textsuperscript{1,2,*}\orcidlink{0009-0006-3937-2954}, 
Adel HAFIANE\textsuperscript{1}\orcidlink{0000-0003-3185-9996}, 
Laurent BOBELIN\textsuperscript{2}\orcidlink{0000-0002-3268-4203}, 
Raphael CANALS\textsuperscript{3}\orcidlink{0000-0001-9100-7539} \\
\IEEEauthorblockA{\textsuperscript{1}INSA CVL, Université d'Orléans, PRISME Laboratory, EA 4229, Bourges, 18022, France} \\
\IEEEauthorblockA{\textsuperscript{2}INSA CVL, Université d'Orléans, LIFO Laboratory, EA 4229, Bourges, 18022, France} \\
\IEEEauthorblockA{\textsuperscript{3}Université d'Orléans, INSA CVL, PRISME Laboratory, EA 4229, Orléans, 45072, France} \\
\thanks{\textsuperscript{*}Corresponding author's email: denis.mamba\_kabala@insa-cvl.fr (Denis MAMBA KABALA \orcidlink{0009-0006-3937-2954})}
}

\maketitle

\begin{abstract}
        Crop disease detection and classification is a critical challenge in agriculture, with major implications for productivity, food security, and environmental sustainability. While deep learning models such as CNN and ViT have shown excellent performance in classifying plant diseases from images, their large-scale deployment is often limited by data privacy concerns. Federated Learning (FL) addresses this issue, but centralized FL remains vulnerable to single-point failures and scalability limits. In this paper, we introduce a novel Decentralized Federated Learning (DFL) framework that uses validation loss ($Loss_{val}$) both to guide model sharing between peers and to correct local training via an adaptive loss function controlled by weighting parameter. We conduct extensive experiments using PlantVillage datasets with three deep learning architectures (ResNet50, VGG16, and ViT\_B16), analyzing the impact of weighting parameter, the number of shared models, the number of clients, and the use of $Loss_{val}$ versus $Loss_{train}$ of other clients. Results demonstrate that our DFL approach not only improves accuracy and convergence speed, but also ensures better generalization and robustness across heterogeneous data environments making it particularly well-suited for privacy-preserving agricultural applications.    
\end{abstract}


\section{Introduction}

Agriculture plays a crucial role in economic, social and environmental development, while ensuring global food security. The sector employs over 27\% of the world's workforce and contributes 4\% to global GDP \cite{Fao2023yearbook}. However, crop diseases represent a serious threat to agricultural productivity, causing yield losses estimated at between 20\% and 40\% per year \cite{Hossain2024}. These losses compromise not only farmers' incomes, but also global food security. In addition, untreated plant diseases can introduce toxic substances into the food chain, posing risks to human health \cite{Aktar2009}, \cite{Savary2019}. In this context, artificial intelligence (AI) tools stand out as promising solutions for early and accurate disease detection, contributing to sustainable agricultural practices \cite{Bhargava2024}, \cite{Javidan2024}, \cite{Ali2024}, \cite{Chaudhari2024}, \cite{Sajitha2024}, \cite{Li2021}.

Recent advances in computer vision have made it possible to automate many agricultural tasks, including the detection and classification of crop diseases \cite{Tejaswini2024}, \cite{Attri2023}. Traditional diagnostic approaches, often time-consuming and costly, are gradually being replaced by machine learning (ML) and deep learning (DL) techniques. Advanced models, such as convolutional neural networks (CNNs) \cite{Tugrul2022}, \cite{Hassan2021},  vision transformers (ViTs) \cite{Singh2024}, \cite{Borhani2022}, \cite{Vallabhajosyula2024}, or CNN and ViT simultaneously \cite{Yang2024}, \cite{Zeng2020}, have proved highly effective in processing complex data. These tools offer high accuracy for classifying crop diseases based on image analysis \cite{Demilie2024}, \cite{Omaye2024}, thus enhancing agricultural quality and productivity . However, their large-scale deployment is hampered by challenges linked to the collection of consistent data in large quantities and the use of such data, not least because of confidentiality issues.

Deep learning models typically rely on large amounts of data, which poses major challenges in terms of privacy, data security and compliance with regulations such as the RGPD. In agriculture, information on production methods, crop disease detection techniques and disease management practices is considered sensitive by economic players, and sharing it represents a competitive risk \cite{Dembani2025}, \cite{alik2023}, \cite{Behera2025}. Federated learning (FL), a machine learning paradigm, offers a viable solution to these challenges. This approach enables several institutions to collaborate in training a global model without the need to share private local data \cite{Wen2022}, \cite{Zhang2021}. Only the parameters of the participants' various local models are sent to a central server to be aggregated to form the global model, which is then returned to all participants \cite{Guendouzi2023}, \cite{MambaKabala2023}.  By enabling collaborative model formation without direct sharing of private data with a central server, federated learning reduces data transfer costs while preserving confidentiality \cite{Mothukuri2021}. In addition, by minimizing the sharing of personal data and giving users control over their own data, federated learning reduces the risk of leakage of sensitive data. While centralized FL, with its central server, is widely adopted in various fields \cite{yang2019}, it remains vulnerable to high communication costs and single point of failure (server) failures. These limitations make the centralized approach ill-suited to agricultural contexts.

In this context, decentralized federated learning (DFL) offers a robust alternative by removing dependency on a central server and promoting peer-to-peer communications \cite{Zhou2024}, \cite{Maenpaa2021}, \cite{Wang2023}. DFL has established itself as an essential approach to machine learning, enabling collaborative model learning by multiple clients without dependence on a central server \cite{Yuan2023}, \cite{Chai2024}. This architecture improves accessibility and resilience, particularly in rural areas with irregular connectivity. In our framework, we have developed an advanced strategy based on validation loss ($Loss_{val}$) to not only guide model sharing between neighbors, but also correct local learning. Model sharing is carried out selectively: each client compares its own $Loss_{val}$ with that of its neighbors. If a neighbor's $Loss_{val}$ is lower, that neighbor is considered to have a better model, and its model is then shared with the other clients. This ensures that only the best-performing models circulate in the network, improving overall convergence. In addition to sharing, the $Loss_{val}$ is also used to adjust the local learning process. When a customer receives a better model, it records the $Loss_{val}$ associated with that model and incorporates it into the calculation of its local loss function. This approach allows local models to be corrected in line with global performance, while limiting the influence of external models through adaptive control via $\lambda$. This makes the system both robust and flexible, suitable for a variety of scenarios and environments.

In this paper, we present a new approach combining $Loss_{val}$-based learning sharing and correction in a decentralized federated learning framework applied to crop disease classification. Our experiments, performed on datasets from PlantVillage \cite{data}, demonstrate that this strategy improves convergence speed, reduces communication costs and increases model accuracy. The remainder of this article is organized as follows: section 2 reviews related work, section 3 presents our method and the tools used, section 4 analyzes experimental results, and section 5 concludes by proposing perspectives for future research.

\section{Related work} 
\subsection{Detection of crop diseases using deep learning models}
Image-based crop disease detection has become a major application of deep learning models, offering promising results in terms of accuracy and robustness. Numerous recent works relevant to image-based crop disease classification tasks show a particular focus on CNN architectures \cite{Mukherjee2025}, \cite{Sajitha2024}, \cite{Ali2024}, \cite{Rahman2025}, \cite{Tejaswini2024}, \cite{Omaye2024}, \cite{Javidan2024}, due to their resource-efficiency and speed, with pre-trained models retaining an advantage in terms of accuracy and stability \cite{Demilie2024}, \cite{Hassan2021}. Recent work \cite{Singh2024}, \cite{Vallabhajosyula2024}, \cite{Yang2024} has also demonstrated the strong potential of ViT models for complex, high-resolution tasks, offering improved contextual understanding, particularly useful for fine classification and accurate localization of crop diseases.

In \cite{Sajitha2024}, the authors performed a synthesis of Machine Learning (ML) and Deep Learning (DL) techniques for crop disease classification in industrial agricultural systems. The study recommends CNNs for real-world deployments. In \cite{Demilie2024} and \cite{Omaye2024}, the authors conducted comparative studies between several image-based crop disease detection and classification techniques. Pre-trained CNNs are identified as the most efficient, with accuracies above 95\%, underlining their effectiveness. In \cite{Mukherjee2025}, the authors proposed a comprehensive review of rice disease identification methods using machine learning techniques. The study highlights the effectiveness of CNN models such as VGG16, ResNet50 and InceptionV3, which outperform conventional methods, with accuracies of up to 98\%. A review focusing on engineering techniques combining CNNs and classical artificial intelligence methods is offered in \cite{Javidan2024}. The authors point out that these approaches offer greater robustness and adaptability to field conditions. In \cite{Ali2024}, the authors developed a CNN model ensemble method (VGG19, DenseNet201, InceptionResNetV2) achieving 99.5\% accuracy, to capture complementary image features, thus improving diagnostic robustness. A real-time monitoring system based on a customized CNN model is developed in \cite{Rahman2025}. The model achieved 97.6\% accuracy, demonstrating the capability of CNNs in live agricultural monitoring. A three-layer convolutional CNN for early disease detection is implemented in \cite{Tejaswini2024}. The model achieves accuracy above 95\%, while maintaining low computational complexity. A combination of deep CNNs with transfer learning (VGG16, InceptionV3, MobileNet) for plant leaf disease classification is proposed in \cite{Hassan2021}. The best model achieves 99.4\% accuracy, proving that pre-trained models are highly effective for plant disease recognition. In \cite{Singh2024}, the authors proposed a model based on pre-trained ViT. The model achieved an accuracy of 97.8\%, demonstrating the ability of ViTs to exploit global representations. A new approach combining residual blocks (ResNet) and visual transformers is introduced in \cite{Vallabhajosyula2024}. The model achieves 98.9\% accuracy, outperforming standard CNNs.

Overall, the literature highlights the dominance of CNN-based models in crop disease detection due to their efficiency, ease of deployment and consistently high accuracy, particularly when using transfer learning. However, emerging vision transformer (ViT) architectures show strong potential in handling complex classification tasks by capturing richer contextual information, particularly useful for good disease localization.

\subsection{Federated Learning (FL)}
FL is a machine learning approach that enables multiple entities to collaborate on model training without sharing their raw data, thus preserving confidentiality and information security. Numerous recent studies on FL have shown it to be a promising solution, offering innovative approaches, reflecting its ongoing evolution and potential for application in a variety of fields, These include agriculture \cite{MambaKabala2023}, \cite{Dembani2025}, \cite{alik2023}, \cite{Idoje2023}, \cite{Behera2025}, medicine \cite{Sharma2023}, \cite{Mishra2023}, \cite{Guan2024}, security \cite{Schoenpflug2024}, IoT \cite{Alsharif2024}, \cite{Belenguer2025}, optimization \cite{Guendouzi2023} and aggregation techniques \cite{Abdelmoula2024}.

In the agricultural domain, FL is essential for improving farm management and smart farming, enabling advanced analysis of agricultural data while guaranteeing security and preserving data confidentiality. \cite{Dembani2025} propose an in-depth study of the different variants of FL applied to agriculture. The study identifies technical challenges and highlights the potential of FL to improve the analysis of agricultural data while preserving confidentiality. \cite{alik2023} provides an overview of FL techniques used in agriculture and the progress made in this field. \cite{MambaKabala2023} uses federated learning to classify leaf diseases of various crops (Grape, Apple, Corn and Tomato) using CNNs and ViTs. The article \cite{Idoje2023} explores the use of FL for crop classification in a decentralized network of smart farms based on local data. According to the results obtained, federally trained models achieve faster convergence and higher accuracy than centralized network models. The integration of deep learning models in an FL environment for crop disease prediction was carried out by the authors in \cite{Behera2025}. The proposed FL approach enabled early and accurate detection of crop diseases, while ensuring data confidentiality and contributing to agricultural sustainability. In the healthcare field, recent work on FL presents significant advances, particularly in addressing confidentiality and collaboration issues in medical image analysis. In \cite{Sharma2023} and \cite{Mishra2023}, the authors carry out an in-depth analysis of the various FL architectures applied to medical data, and identify the technical challenges associated with its application in the medical field. An exhaustive review of FL methods applied to medical image analysis is carried out in \cite{Guan2024}. The review highlights recent advances in FL for medical image analysis, while discussing current challenges and future research opportunities. In the field of security, a study on the use of FL in intrusion detection systems is carried out in \cite{Schoenpflug2024}. The study shows that FL can improve intrusion detection while preserving data confidentiality. In \cite{Alsharif2024}, the authors examine the application of FL at the network edge in the context of the Internet of Things (IoT), focusing on decentralized architectures and learning models adapted to edge devices. Recent FL methodologies and persistent challenges, through detailed taxonomies and a review of various FL applications, particularly in the IoT, are also explored in \cite{Belenguer2025}. \cite{Teixeira2025} analyze the costs associated with training models in FL, comparing different architectures and optimization strategies to improve energy and computational efficiency. In \cite{Abdelmoula2024}, a comprehensive review of the different categories of LF, focusing on inherent challenges, aggregation techniques such as FedAvg and associated development tools, is carried out. The costs associated with training models in FL are analyzed in \cite{Guendouzi2023}. The authors compare different architectures and optimization strategies to improve energy and computational efficiency, while maintaining high model performance.

FL has established itself as a versatile, privacy-friendly paradigm that is highly applicable in a variety of fields, including agriculture. Its ability to enable collaborative model learning without compromising data confidentiality makes it particularly suited to sensitive domains. The various recent works above confirm its relevance not only for smart agriculture and crop disease detection, but also for healthcare, IoT and cybersecurity. The ongoing evolution of FL architectures, optimization strategies and aggregation techniques reinforces its potential as a scalable and secure alternative to centralized learning.

\subsection{Decentralized Federated Learning (DFL)}
DFL is a collaborative approach to training machine learning models without relying on a central server. It has emerged as an essential approach to machine learning, enabling collaborative model training by multiple clients using Peer-to-Peer (P2P) communication, without relying on a central server. Recent studies on this approach illustrate current advances and challenges, focusing on robustness to attacks, communication efficiency, as well as specific applications in various fields. Yuan and al. \cite{Yuan2023} offer an in-depth study of various decentralized federated learning methodologies, highlighting central serverless architectures and client communication topologies. The authors summarize current DFL challenges and suggest future research directions. \cite{Chai2024} introduce AdFed, a DFL method for cancer survival prediction, with a focus on privacy preservation. The proposed method, AdFed, demonstrated superior performance in predicting cancer survival compared to traditional federated methods. In \cite{Toofanee2023}, the authors compare centralized and P2P FL approaches using a Siamese model for the classification of diabetic foot ulcers. According to the results obtained, the P2P approach showed comparable performance to the centralized approach, with better confidentiality preservation. \cite{Salmeron2023} evaluated different aggregation strategies in a decentralized P2P environment applied to biomedical data. Accuracy-based weighted aggregation outperformed the classical federated aggregation method, highlighting the importance of adapted strategies in decentralized environments to improve performance. In \cite{Zhang2024}, the authors propose the integration of blockchain technology into DFL to ensure the security and traceability of model updates. \cite{Maenpaa2021} explores DFL algorithms and compares them with centralized approaches, highlighting the challenges and benefits of DFL, particularly in terms of resilience to failures. \cite{Khan2025} study hyperparameter optimization in a DFL framework, using consensus mechanisms to improve model convergence. The proposed strategies led to faster and more stable model convergence. In \cite{Zhou2024} and \cite{Wang2023}, the authors respectively propose DeFTA and SparSFA, P2P DFL frameworks aimed at facilitating and improving direct communication between clients, reducing communication costs and demonstrating increased resilience in DFL environments.

DFL presents a promising evolution of traditional federated learning by eliminating reliance on central servers and enhancing privacy, scalability, and resilience. The literature shows that peer-to-peer architectures can achieve comparable or even superior performance to centralized methods, especially when supported by adaptive aggregation strategies and robust communication frameworks. Despite ongoing challenges, particularly around coordination and convergence, recent advances confirm the potential of DFL for sensitive and distributed applications such as healthcare, biomedical analysis, and privacy-aware prediction systems.

On the basis of the various works presented, it is clear that deep learning models, in particular CNNs and ViTs, are becoming increasingly successful in the detection of plant diseases. Although FL has emerged as a promising solution for guaranteeing data confidentiality and enabling collaboration between institutions during the global model training process, limitations persist, notably dependence on a central server, introducing single-point-of-failure risks and scalability challenges. DFL has been proposed as an alternative to address the limitations of FL, removing dependence on a central server while enabling P2P collaboration between participants. However, a key challenge that persists in DFL is the strategy for model sharing among participants. In the absence of a central aggregator, the question of how to exchange models in an intelligent and effective manner becomes crucial. Specifically, determining which models should be shared, with whom, and under what conditions, remains an open research problem. The objective is to ensure that each participant  shared model contribute meaningfully to the performance improvement of local learners. In this context, our contribution presents a new DFL framework using a strategy of model sharing between participants and local learning correction, while leveraging the advantages of P2P communication. Starting from this new framework, our study will address questions related to regulating the impact of external data in the local training process via the $\lambda$ parameter on model performance, the impact of the number of models shared between clients on model performance, the impact of varying the number of local devices on model performance, and the impact of using loss on the validation or training data subset during the local learning correction process on model performance. We will study these different questions through experiments on a plant disease classification problem, and using CNN and ViT pre-trained models, in order to provide appropriate answers.

\section{Methodology}
In this section, we present our proposed approach in detail, then review the different deep learning models used in our experiments.
\subsection{Proposed approach}
\begin{figure*}[t]
  \centering
  \includegraphics[width=0.74\textwidth]{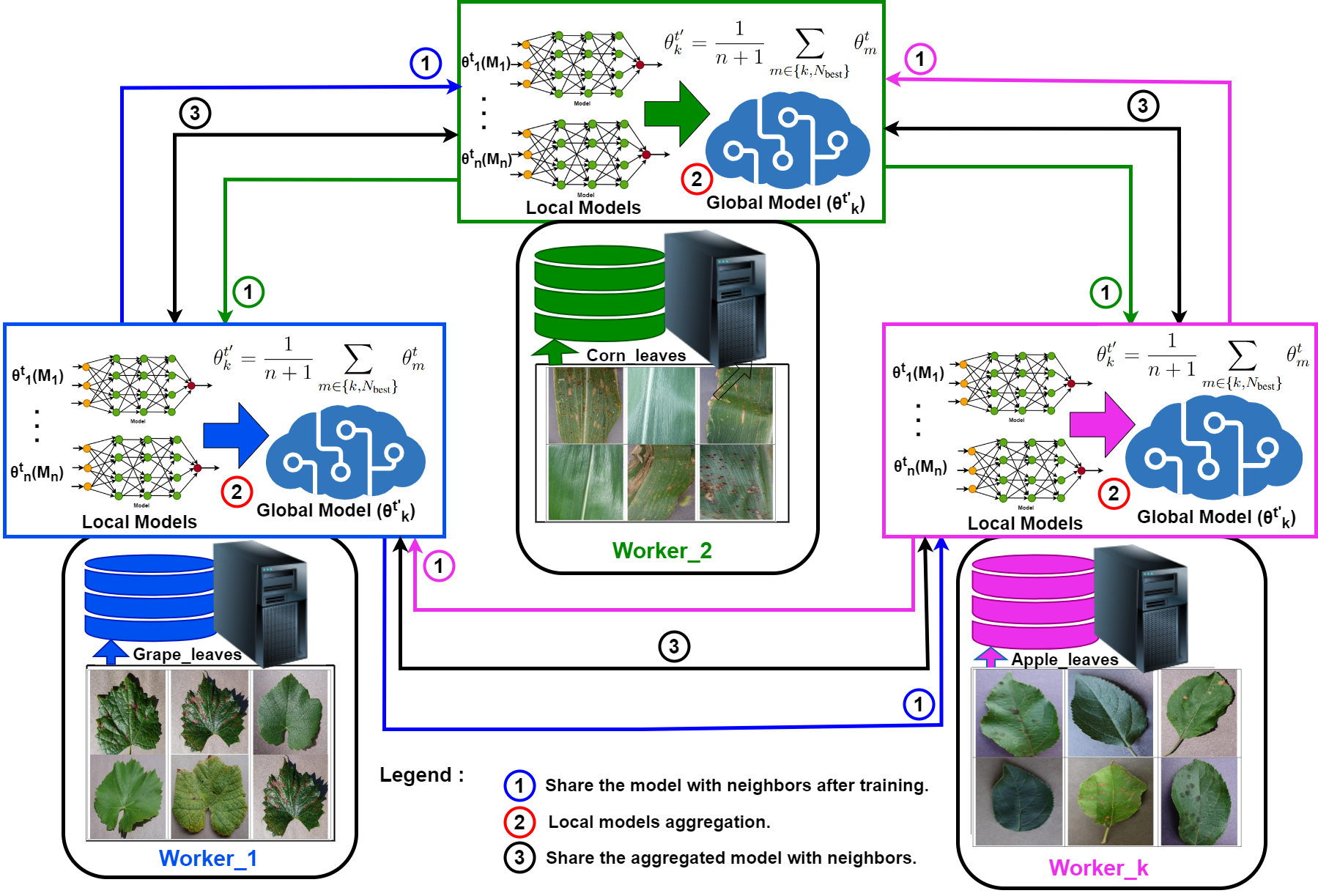}
  \caption{Architecture of the proposed approach.}
  \label{fig:architecture}
\end{figure*}

The proposed approach, as presented in Figure~\ref{fig:architecture}, is based primarily on a decentralized federated learning framework. This is based on peer-to-peer communication between participants for model sharing, without recourse to a central server. The framework comprises several workers or clients, each with its own private data and a pre-trained initial model. Each worker is directly connected to its neighbors, with the aim of enabling collaboration towards the formation of a robust and efficient global model \cite{yang2019}, while keeping private data local and bypassing a central server to orchestrate the process. We simulated a collaborative set of clients or participants contributing to the improvement of a global model. Figure~\ref{fig:architecture} shows the sequence of the different phases of our proposed DFL framework.

Presented in Figure~\ref{fig:architecture}, the proposed approach is structured around five key phases: initialization, local training, models sharing, models aggregation, and update local loss function.
\begin{enumerate}
    \item \textbf{Initialization} :\\    
Let ${C}$ be the number of clients participating in the decentralized federated learning process and ${R}$ the total number of communication cycles or rounds. Each client $k \in \{1,2,\dots,C\}$ predefined in the environment has a pre-trained model $M_{k}^{0}$. Thus, all clients are initialized with the same model \( M_k^0 \). Otherwise, \( \forall (k,j) \in \{1,2,\dots,C\}, M_k^0 = M_j^0 \). The number of participants \( C \) and communication cycles \( R \) is set in this phase.
 	   
    \item \textbf{Local training} :\\ 
Let \( D_k \) be client \( k \)'s local dataset, where \( \forall i,j \in \{1,\dots,C\}, i \neq j \Rightarrow D_i \cap D_j = \emptyset \). In this phase, each client \( k \) performs local learning on its own private data using its initial model \( M_k^0 \).  
Thus, the local model \( L_k^t \) of each client \( k \), at each cycle \( t \), will be updated as follows :
\begin{equation}
    L_k^t = \text{Training} \left( A_k^{t-1}, D_k \right)
    \label{eq:training}
\end{equation}
where \( A_k^{t-1} \) is the global model of client \( k \) at iteration or round \( t-1 \), with $t \in \{1,2,\dots,R\}$. Initially \( A_k^0 = M_k^0 \).
With the three plant leaf datasets at our disposal, we experimented with two configurations: one with six customers (due to the duplication of one dataset for two customers) and the other with 18 customers (due to the duplication of one dataset for six customers), while keeping the size of the datasets identical for each configuration (i.e. 80\% of data for the training subset and 20\% for the test subset). Namely, for each configuration, the training data subset (80\%) is partitioned by the number of clients participating in the process, then each portion is allocated to the corresponding customer for local model training. In other words : 
\begin{equation}
\begin{split}
    \forall (i \neq j); \quad i,j \in \{1,\dots,C\}, \\
    D_i \cap D_j = \emptyset, \\
    D_{\text{train}} = \bigcup_{k=1}^{C} D_{k\_train}.
\end{split}
\end{equation}
The idea behind this experiment is to study the impact of varying the number of participants in the FL process on the performance of the overall model. In this phase, model performance evaluation measures are calculated for each participant \( k \) at each communication cycle \( t \), and the local validation loss is estimated as follows:
\begin{equation}
    \text{Loss}_{\text{val},k}^t
 = \text{Loss} \left( A_k^t,\ D_{k\_test} \right)
    \label{eq:val_loss}
\end{equation}
Where $A_k^t$ is the model of client \( k \) at round \( t \), and $D_k^{test}$ is the test or validation data partition of the same client.

    \item \textbf{Models sharing} :\\
Let \( N_k \) be the set of neighbors of client \( k \), with \( |N_k| = C-1 \). 
At each cycle \( t \), \( k \) shares its model with its satisfying \( N_k \) neighbors : 
\( \forall j \in N_k, \quad j \neq k, \quad \text{if } \text{Loss}_{\text{val},j}^t > \text{Loss}_{\text{val},k}^t \) 
then \( k \) shares its model with \( j \).
Our idea behind the above criterion is to allow each participant to share (send or receive) model parameters directly with all its neighbors who satisfy criterion (3), and thus enable the best performers in each cycle to contribute to improving the performance of their neighbors, thus promoting learning and rapid convergence of the overall model. Based on our two client configurations (six and 18), we will experiment with two scenarios: the case of sharing a single best model and that of sharing five best models in each cycle. The aim is to study the impact of the number of models shared between participants, to see to what extent the communication load can be reduced while achieving better performance for our overall model. By analyzing the results obtained for each of our two configurations, we will also be able to determine which of the two scenarios is best suited to the proposed new framework.

    \item \textbf{Models aggregation} :\\
Let \( \theta_k^t \) be the model parameters of each participant \( k \) at cycle \( t \). The aim of this phase is to allow, for each participant \( k \) and at each cycle \( t \), to locally aggregate the customer's private model as well as the \( n \) best models received from \( N_k \) neighbors using the FedAvg algorithm. This algorithm aggregates the local models by averaging the weights to obtain a global model. Thus, the aggregated model \( \theta_k^{t'} \) for each participant \( k \) , at each cycle \( t \) , will be estimated as follows : 
\begin{equation}
    \theta_k^{t'} = \frac{1}{\lvert N_{best} \rvert +1} \sum_{m \in \{k, N_{\text{best}}\}} \theta_m^t
    \label{eq:aggregation}
\end{equation}
where \( N_{\text{best}} \) is the set of the best models received by a given client $k$, and $\lvert N_{best} \rvert$ is the cardinal of $N_{\text{best}}$.  
At the end of this phase, each participant will have his own local global model, obtained by aggregating his local model \( L_k^t \) and the \( N_{\text{best}} \) models received from his neighbors.
 
    \item \textbf{Update local loss function} :\\
At the end of phase 3, when a customer receives a better model, after performing aggregation as explained in phase 4, it adjusts its local loss function by integrating the loss value on the test data subset ($Loss_{val}$) of this better model received. The new local loss function is defined by :
\begin{equation}
    \text{Loss}_k^t = \mathrm{Local\_loss}_k^t + \lambda \cdot \text{Loss}_{\text{received\_model}}^t
    \label{eq:new_loss}
\end{equation}
Where :
\begin{itemize}
    \item \( \mathrm{Local\_loss}_k^t \) is the loss of the local model on its own subset of training data.
    \item \( \text{Loss}_{\text{received\_model}}^t \) is the loss on the validation data subset of the best model received.
    \item \( \lambda \) is an experimental hyperparameter for dynamically controlling or adjusting the influence of shared (or external) models on local learning.
\end{itemize}
If participant \( k \) receives \( n \) best models, then \( \text{Loss}_{\text{received\_model}}^t \) will be the average loss of all \( n \) best models received. It is calculated as follows :

\begin{equation}
    \text{Loss}_{\text{received\_model}}^t = \frac{1}{\lvert N_{best} \rvert} \sum_{m \in N_{\text{best}}} \text{Loss}_{\text{val},m}^t
    \label{eq:received_loss}
\end{equation}
The new value of the loss function obtained will be used in the next cycle in the local learning of the model to correct or readjust the learning, in order to speed up the convergence of the global model while making as few communications as possible. As an experimental hyperparameter, we empirically chose and tested four specific values of $\lambda \in [0, 1]$, namely : 0, 0.25, 0.50 and 0.75, in order to assess their impact on model performance and to determine, after an analysis of the results obtained for each configuration, which of the $\lambda$ values is optimal in our case. This approach makes it possible to regularize local learning by taking global performance into account. According to (5), we have also experimented, for our two configurations, with scenarios in which we use the loss on the training data subset ($Loss_{train}$) of the best model, instead of using $Loss_{val}$ to calculate the new loss function. In this context, (6) becomes :

\begin{equation}
    \text{Loss}_{\text{received\_model}}^t = \frac{1}{\lvert N_{best} \rvert} \sum_{m \in N_{\text{best}}} \text{Loss}_{\text{train},m}^t
    \label{eq:train_loss} 
\end{equation}
Our aim in (7) is to determine which of the two losses (val\_loss or train\_loss) provides the best performance optimization after the learning correction using the new loss function. Once the new value of the loss function has been calculated in both cases, the process is repeated from phase 2 for the next cycle.
\end{enumerate}
This five-phase cycle is repeated until the predefined number of communication cycles between participants is reached. It is therefore clear that, in this new approach, each participant enables, on the one hand, the local training of models, and on the other, the aggregation of the different models received from its neighbors, as well as the estimation of the new loss function, the value of which will be used in the next cycle to adjust learning. Unlike centralized approaches, we don't depend on a central server to orchestrate the process. Each participant is autonomous, which offers greater guarantees of data confidentiality.

The following algorithm summarizes all five process phases of our proposed new approach described above.

\begin{algorithm}[H]
\caption{: Decentralized FL with Val\_loss for model sharing and local learning correction}\label{alg:hsfedl2}
\begin{algorithmic}[1]
\State \textbf{Data}: Each client $k$ holds a local dataset $D_k$.
\State \textbf{Hyperparameters}: $E$ is the number of local epochs; $B$ is the mini-batch size; $\eta$ is the learning rate; $\lambda$ is the weight parameter of the external val\_loss; $R$ is the number of communication rounds.
\State \textbf{Initialization}:
\For{each client $k$}
    \State Initialize model $A_k^{0} \gets M_k^0$
\EndFor
\For{each round $t$ from 1 to $R$}
    \For{each client $k$ \textbf{in parallel}}
        \State Split $D_k$ into mini-batches $B$;
        \For{each local epoch $j$ from 1 to $E$}
            \For{each batch $b \in B$}
                \State $L_k^t \gets L_k^t - \eta\nabla L(A_k^{t-1}; b)$;
            \EndFor   
        \EndFor
        \State $Loss\_val_k^t = \text{Loss}(A_k^t, D_{k\_test})$; \Comment{Validation loss computation.}
    \EndFor
    \For{each client $k$ \textbf{in parallel}}
        \State Recover $Loss\_val_j^t$ from all neighbors $N_j$ of $k$;
        \State Select the $n$ best-performing models (lowest $Loss_{val}$);
        \State Share $L_k^t$ with neighbors having higher $Loss_{val}$;
        \State Collect received models $L_{N_{best}}^t$;
        \State $A_k^t \gets \text{FedAvg}(L_k^t, L_{N_{best}}^t)$; \Comment{Model aggregation.}
    \EndFor
    \For{each client $k$ \textbf{in parallel}}
        \State Compute adjusted loss :
            \State $Loss_{received\_model}^t \gets \frac{1}{\lvert N_{best} \rvert} \sum_{m \in N_{best}} Loss\_val_m^t$

            \State $Loss_k^t = \mathrm{Local\_loss}^{t}_{k} + \lambda \cdot Loss_{received\_model}^t$;
    \EndFor
\EndFor
\State \textbf{End of process.}
\end{algorithmic}
\end{algorithm}

\subsection{Deep learning architectures}
In our experiments, we implemented tree pre-trained models, chosen for their widespread use in the literature and their proven performance in image classification tasks. These models include two convolutional neural networks (CNNs) and one vision transformer (ViT), offering a range of strengths and trade-offs that are particularly relevant for agricultural applications \cite{Attri2023}. Below, a brief overview of these architectures, emphasizing their strengths and weaknesses in the context of image classification for crop disease detection.

\begin{itemize} 
\item \textbf{ResNet50 :} \cite{he2015deep} is a 50-layer convolutional neural network built on the VGG architecture. Its innovative residual blocks address the vanishing gradient problem by enabling gradients to propagate effectively through deeper layers, allowing for the training of very deep networks. ResNet50 excels in image classification due to its strong feature extraction capabilities, which are essential for detecting fine-grained differences in crop diseases. However, its computational requirements pose a problem of adaptability when deployed on devices with limited resources.

\item \textbf{VGG-16 :} Developed by the Visual Geometry Group at Oxford\cite{simonyan2015deep}, is a 16-layer deep learning model that builds upon AlexNet. By using smaller 3×3 convolution filters, it reduces the number of parameters while maintaining high accuracy. Its simplicity and effectiveness make it a strong choice for image classification tasks. However, its memory costs may limit its scalability for large data sets.

\item \textbf{Vision Transformer (ViT):}  
The Vision Transformer \cite{dosovitskiy2021image}, introduced in 2021, adapts the transformer architecture originally designed for natural language processing to image classification. By dividing images into patches and applying self-attention mechanisms, ViT excels at capturing global dependencies in images. This makes it highly effective for recognizing complex patterns in crop diseases. However, ViTs typically require large training datasets and are computationally intensive, which can be a limitation in agricultural scenarios with limited resources or smaller datasets.
\end{itemize}

\section{Experiments and results}\label{sec4} 
In this section, we describe the various experiments carried out as part of our study. We detail the experimental configuration of our development environment, the different measures used to evaluate the performance of our models, and the results obtained and their interpretation for each experimental scenario.

\subsection{Experimental procedure}

\subsubsection{Description of the datasets}
\begin{figure*}[t]
    \centering
    \includegraphics[width=\linewidth]{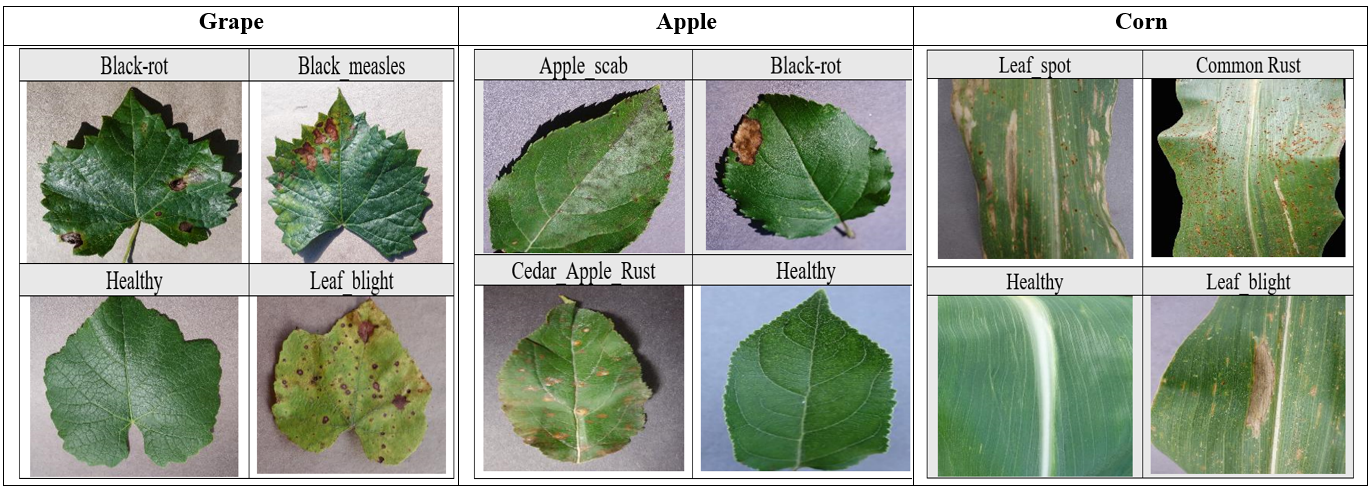}
    \caption{Example of images from our datasets.}
    \label{fig:Images}
\end{figure*}

To train and evaluate our federated learning framework, we used public data from the PlantVillage database \cite{data}. This database contains over 50,000 images of plant leaves, both healthy and infected, classified into 38 categories according to species and pathologies. It constitutes a freely available image resource on plant health, useful for the development of disease diagnostics. In our study, we selected data related to three types of plant for our experiments: grapes, apples and corn. These data are organized into different categories of healthy and diseased leaves. Table~\ref{tab:dataset} provides a detailed description of these categories. Figure~\ref{fig:Images} shows some images of leaves belonging to the different classes for each of the three plants.

\begin{table*}[t]
\centering
\caption{Datasets description.}
\label{tab:dataset}
\begin{tabular}{|l|l|c|c|}
\hline
\textbf{Datasets name}          & \textbf{Class name} & \textbf{Number of images/class} & \textbf{Number of images/dataset} \\ \hline
\multirow{4}{*}{\textbf{Grape}} & Black\_rot          & 1180                            & \multirow{4}{*}{\textbf{4062}}    \\
                                & Black\_Measles      & 1383                            &                                   \\ 
                                & Leaf\_blight        & 1076                            &                                   \\ 
                                & Healthy            & 423                             &                                   \\ \hline
\multirow{4}{*}{\textbf{Apple}} & Apple\_scab         & 630                             & \multirow{4}{*}{\textbf{3171}}    \\ 
                                & Black\_rot          & 621                             &                                   \\ 
                                & Cedar\_apple\_rust  & 275                             &                                   \\ 
                                & Healthy             & 1645                            &                                   \\ \hline
\multirow{4}{*}{\textbf{Corn}}  & Leaf\_spot          & 513                             & \multirow{4}{*}{\textbf{3852}}    \\ 
                                & Common\_rust        & 1192                            &                                   \\ 
                                & Healthy            & 1162                            &                                   \\ 
                                & Leaf\_blight        & 985                             &                                   \\ \hline
\end{tabular}
\end{table*}

\subsubsection{Hardware setup}
The experiments were carried out using Python 3.9.12, on a computer equipped with an Intel(R) Xeon(R) W-2223 processor and an NVIDIA Quadro RTX 5000 graphics card, running Ubuntu 20.04.4 LTS 64-bit. We used the PyTorch 1.10.2 library to develop deep learning architectures, as well as cuDNN 8.2.1 and the CUDA 11.2.67 toolkit to run the algorithms on the GPU. We ran the experiments on three datasets, using the parameters defined in Table~\ref{tab:params}.
\begin{table*}[t]
\centering
\caption{Experimental parameters.}
\label{tab:params}
\begin{tabular}{|c|c|c|c|c|c|c|}
\hline
 & \textbf{Client} & \textbf{Epoch/Client} & \textbf{Batch Size} & \textbf{Rounds} & \textbf{Classes} & \textbf{Learning rate} \\ 
\hline
\textbf{Values} & 6, 18 & 1 & 16 & 50 & 4 & 0.01 \\ 
\hline
\end{tabular}
\end{table*}

\subsubsection{Configuration of learning clients}
For the experimental setup, we used three datasets, described in Table~\ref{tab:dataset}, with an 80\%-20\% partitioning strategy, for model training and testing respectively. We designed two client configurations : one with 6 clients and the other with 18 clients. In the 6-client configuration, we duplicated each of the three data sets to two (2) clients, subdividing the 80\% of the training data partition equally between the two (2) clients, while keeping the 20\% of the test data partition identical in size for each of the two (2). On the other hand, in the 18-client configuration, we duplicated each of the three data sets to six (6) clients, due to a subdivision of the 80\% of the training data partition equally between the six (6) clients, while keeping the 20\% of the test data partition identical in size for each of the six (6). For each of our two client configurations mentioned above, we used a 50 communication cycle between clients, one (1) epoch locally for each client participating in the process, a batch size of 16 and a learning rate set at 0.01. Our aim behind these experiments, through these two configurations, is to study the impact of varying the number of customers on model performance, the impact of the number of shared models on model performance, the impact of the size of the training data subset on model performance, the influence of the hyperparameter $\lambda$ on model performance as a dynamic factor for adjusting the impact of external models on local learning, and finally the study of the impact of using $Loss_{val}$ or $Loss_{train}$ in the calculation of the new loss function on model performance. These different aspects ensure that the proposed method is tested in different configurations, providing a better understanding of its robustness and adaptability to various scenarios.

\subsubsection{Performance evaluation measures}
We used four measures to evaluate the performance of our different models, namely precision, recall, accuracy and F1 score. These measures are key to understanding the strengths and weaknesses of our models, and they take into account indicators based on the number of true positives (TP), true negatives (TN), false positives (FP) and false negatives (FN) \cite{Sokolova2009}.

\begin{itemize}
    \item \textbf{Accuracy :} It provides an overall estimate of the proportion of correct predictions in relation to all predictions. It is expressed as follows :
    
    \begin{equation}
    \text{Accuracy} = \frac{TP + TN}{TP + FP + TN + FN} \label{eq:accuracy}
    \end{equation}
    \item \textbf{Precision :} It measures the ability of a model to not classify a negative sample as positive. It is expressed as follows :

    \begin{equation}
    \text{Precision} = \frac{TP}{TP + FP} \label{eq:precision}
    \end{equation}
    \item \textbf{Recall :} It measures the ability of a model to identify all positive samples. It is given as follows : 

    \begin{equation}
    \text{Recall} = \frac{TP}{TP + FN} \label{eq:recall}
    \end{equation}
    \item \textbf{F1-Score :} This is the harmonic mean of precision and recall. It measures the global performance of a model by taking into account both precision and recall, but also by giving a compromise between the two. It is expressed as follows :

    \begin{equation}
    \text{F1-Score} = 2 \times \frac{\text{Recall} \times \text{Precision}}{\text{Recall} + \text{Precision}} \label{eq:f1score}
    \end{equation}
\end{itemize}
As we proceed, we will use only two measures to present the performance of the models studied, namely F1-Score and Accuracy.

\subsection{Results and discussion} 
we present the results of our different models according to the various experimental scenarios with model performance evaluation , namely: the influence of the parameter $\lambda$, the impact
of varying the number of clients, the influence of the number of shared
models, and finally the impact of using $Loss_{val}$ or $Loss_{train}$ in the calculation of the new local loss function. For each of these scenarios, we used the parameters defined in Table~\ref{tab:params}.

\subsubsection{Impact of the $\lambda$ parameter }
In this section, we studied the influence of the hyperparameter $\lambda$ used in the calculation of the new loss function. The calculation of this function is based on simultaneous use of the loss of the local model and that of the model received from neighbors, weighted by the value of $\lambda$ in order to control its impact on local learning while optimizing the performance of the global model. The idea is to find, after an empirical study of the different values of $\lambda$ taken from the interval [0, 1], the one that optimizes the performance of the overall model. Starting with each of our two configurations (6 and 18 customers), we used equations (5) and (6) to calculate the new loss function and tested 4 values of $\lambda$, respectively: 0, 0.25, 0.50 and 0.75. For this study, we will present the results for a configuration of 6 clients, with a scenario of sharing one (1) best model between participants, considering 50 cycles of communication between participants and one epoch (epoch = 1) locally for each participant. Table~\ref{tab:lambdaImpact} shows the mean values and standard deviations of the F1\-score and accuracy for our three architectures as a function of the four $\lambda$ values tested. With three datasets (Grape, Apple, Corn), two clients share 80\% of the training data from each of the three datasets, and the remaining 20\% is used by each for testing the overall model.

\renewcommand{\arraystretch}{1.5}
\setlength{\tabcolsep}{8.5pt}
\begin{table*}[t]
\centering
\caption{Results from the study of the impact of the $\lambda$ parameter in a 6-client configuration, with 1 best shared model.}
\label{tab:lambdaImpact}
\begin{tabular}{|c|cc|cc|cc|}
\hline
\multirow{2}{*}{\textbf{$\lambda$}} & \multicolumn{2}{c|}{\textbf{ResNet50}} & \multicolumn{2}{c|}{\textbf{VGG16}} & \multicolumn{2}{c|}{\textbf{ViT\_B16}} \\ \cline{2-7} 
& \multicolumn{1}{c|}{\textbf{F1-Score}} & \textbf{Accuracy} 
& \multicolumn{1}{c|}{\textbf{F1-Score}} & \textbf{Accuracy} 
& \multicolumn{1}{c|}{\textbf{F1-Score}} & \textbf{Accuracy} \\ \hline

\textbf{0.00} & \multicolumn{1}{c|}{95,85 ± 8,45} & 95,86 ± 8,45 & \multicolumn{1}{c|}{98,20 ± 1,23} & 98,24 ± 1,18 & \multicolumn{1}{c|}{97,08 ± 2,34} & 97,21 ± 2,14 \\ \hline
\textbf{0.25} & \multicolumn{1}{c|}{91,14 ± 11,37} & 90,67 ± 13,95 & \multicolumn{1}{c|}{\textbf{98,91 ± 1,35}} & \textbf{98,76 ± 1,68} & \multicolumn{1}{c|}{\textbf{97,71 ± 1,85}} & \textbf{97,55 ± 2,17} \\ \hline
\textbf{0.50} & \multicolumn{1}{c|}{93,21 ± 8,92} & 92,84 ± 10,33 & \multicolumn{1}{c|}{98,71 ± 0,90} & 98,72 ± 0,88 & \multicolumn{1}{c|}{97,48 ± 2,38} & 97,55 ± 2,28 \\ \hline
\textbf{0.75} & \multicolumn{1}{c|}{\textbf{99,00 ± 0,80}} & \textbf{99,02 ± 0,78} & \multicolumn{1}{c|}{98,50 ± 1,04} & 98,50 ± 1,03 & \multicolumn{1}{c|}{97,37 ± 1,57} & 97,38 ± 1,56 \\ \hline
\end{tabular}
\end{table*}

A general analysis of the results in Table~\ref{tab:lambdaImpact} shows that local use of the $Loss_{val}$ of the best neighbor model, weighted by the $\lambda$ parameter for learning correction, has a positive impact on the performance of our three tested architectures, which achieve better performance. They all achieve maximum average performance in scenarios where $\lambda > 0$, with ResNet50 achieving higher average performance than VGG16 and ViT\_B16, reaching a maximum value of 99\% F1\-score, compared with maximum average performance of 98.91\% and 97.71\% F1-score, respectively for VGG16 and ViT\_B16.

The ResNet50 model is strongly impacted by $\lambda$, with a drop in performance and high variability at $\lambda = 0.25$, which could indicate instability in the local model fitting process. It reaches its peak performance (99\% F1-score and 99.02\% accuracy) and optimal stability at $\lambda = 0.75$, suggesting that increased local integration of the received model's $Loss_{val}$ significantly improves local learning. Consequently, a high value of $\lambda$ in the case of ResNet50 results in better learning performance and stability. Furthermore, when analyzing the stability of ResNet50 between local learning with and without external information, the standard deviation drops from 8.45 (with $\lambda = 0$, i.e. without use of external information in local learning) to 0.80 (with $\lambda = 0. 75$, i.e. with the use of external information in local learning), i.e. a reduction of 90.53\%, proving that ResNet50 benefits most strongly from the use of external $Loss_{val}$ in local learning correction and, consequently, enables better overall model convergence. However, despite outperforming the other models, ResNet50 shows a strong dependence on external information to achieve better performance, implying little consideration of the representativeness of local data in the learning process.

Unlike ResNet50, the VGG16 model is not strongly impacted by $\lambda$. It shows consistently high stability, with average performance greater than or equal to 98.50\% F1-score for values of $\lambda > 0$, slightly below that of ResNet50, and achieves a better score (98.91\% F1-score) for $\lambda = 0.25$, suggesting that a weak local influence of the external $Loss_{val}$ is sufficient for VGG16 to achieve maximum performance. For values of $\lambda = 0.50$ and $\lambda = 0.75$, we observe a drop in performance for VGG16, which could suggest saturation of the impact of the external model during local learning. The standard deviation drops from 1.23 (with $\lambda = 0$, i.e. without the use of external information in local learning) to 0.90 (with $\lambda = 0.75$, i.e. with the use of external information in local learning), i.e. a variability reduction of 26.83\%, with a less effect than for ResNet50, suggesting less sensitivity to the integration of the external $Loss_{val}$ in local learning. Furthermore, although benefiting from the correction of local learning via external $Loss_{val}$, VGG16 takes strong account of the representativeness of local data while guaranteeing constant stability, making it the most robust choice if network stability is a priority.

Like VGG16, ViT\_B16 is also less impacted by $\lambda$, compared with ResNet50. It shows fewer fluctuations and consistent stability as a function of $\lambda$, with slightly lower performance than VGG16, and a peak F1-score of 97.71\% for $\lambda = 0.25$. With $\lambda = 0.75$, there is a slight decrease in performance despite the improvement in stability, suggesting that, like VGG16, ViT\_B16 is also less sensitive to the integration of $Loss_{val}$ in the local learning process, and that a strong external influence locally through high values of $\lambda$ degrades learning and leads, consequently, to a reduction in model performance. We observe a 32.9\% reduction in standard deviation, from 2.34 (with $\lambda = 0$, i.e. without use of external information in local learning) to 1.57 (with $\lambda = 0.75$, i.e. with use of external information in local learning), a reduction also less spectacular than for ResNet50, but slightly greater than for VGG16, suggesting, as for VGG16, less sensitivity to the integration of external $Loss_{val}$ in local learning.

\begin{figure*}[t]
    \centering
    \includegraphics[width=\linewidth]{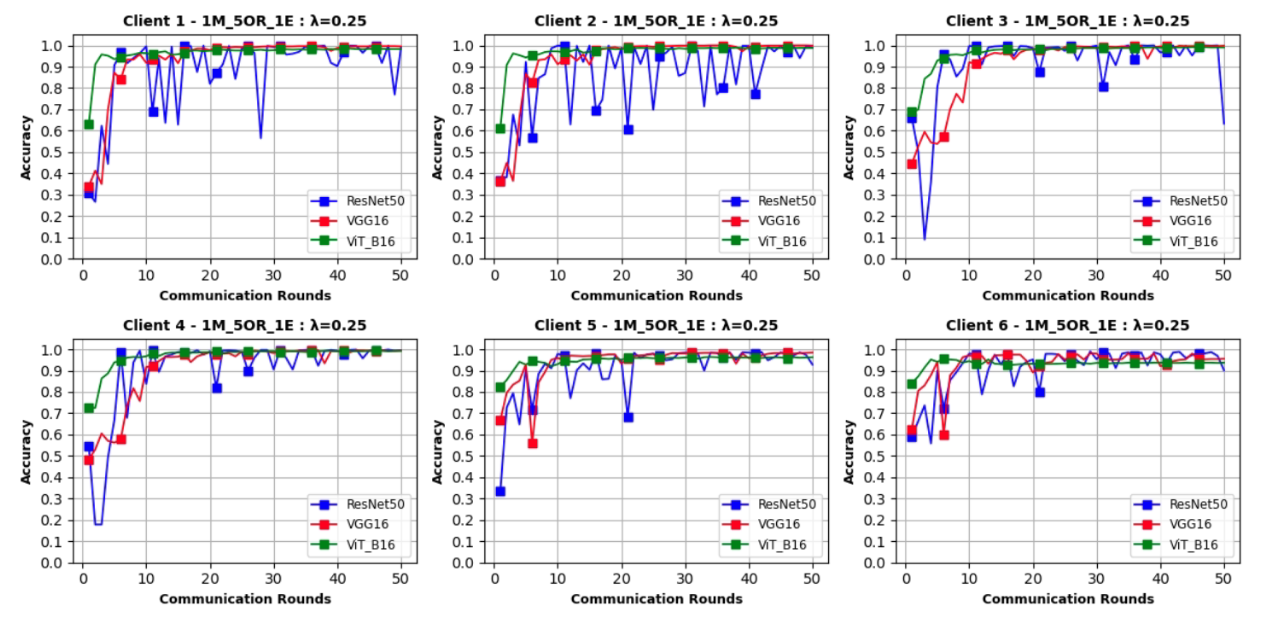}
    \caption{Impact of  the $\lambda$ parameter, in 6-client configuration, one best shared model, with $\lambda = 0.25$.}
    \label{fig:SharedModel1}
\end{figure*}

It's clear that, in the scenario where $\lambda = 0.25$, the VGG16 and ViT\_B16 architectures achieve their best performance and good stability, in contrast to ResNet50, which exhibits high instability. Figure~\ref{fig:SharedModel1} shows the evolution of accuracy in the case where $\lambda = 0.25$ for these three models, according to each of the six clients after 50 communication rounds.

From Figure~\ref{fig:SharedModel1}, we can see an overall improvement trend for all three models. The accuracy of all three models increases progressively over the rounds. The ResNet50 model shows more fluctuations for all clients, with occasional decreases in accuracy over the rounds. This suggests that incorporating the external $Loss_{val}$ weighted by $\lambda = 0.25$ does not help local learning correction and that, consequently, local control via $\lambda = 0.25$ is ill-suited to the local data distribution for ResNet50. In contrast to ResNet50, the VGG16 and ViT\_B16 models show high stability, with a rapid rise in performance for ViT\_B16 and some small oscillations at the start for VGG16. More stable overall, they achieve better performance and converge rapidly as early as 10 rounds for ViT\_B16 and 15 rounds for VGG16, for the majority of customers. This suggests that these two architectures are, in general, an ideal choice in a context where minimizing communication costs, while having less sensitivity to the integration of external information via $Loss_{val}$ (with $\lambda = 0.25$) during local learning and retaining the representativeness of local data, is a priority. However, in a scenario where a compromise between slightly higher communication cost and slightly higher performance is sought, VGG16 would be preferable to ViT\_B16.

In summary, the use of the parameter $\lambda$ in the local learning correction process via the calculation of the new loss function has a significant impact on the performance of our three architectures studied. In scenarios where $\lambda > 0$ (i.e. where external information is used in local learning), we observe a general improvement in the average performance of all the models tested. The ResNet50 model is more impacted by external influence via the external $Loss_{val}$ used when correcting local learning and, as a result, does not take sufficient account of the representativeness of local data in the learning process. In the context of Federated Learning, where network stability and the representativeness of local data in learning are assets, ResNet50 would not be an interesting choice in this framework despite its high performance. VGG16 and ViT\_B16 also benefit from the correction of local learning via external $Loss_{val}$, achieving high performance with fewer communication turns. However, the effect of this benefit is less pronounced compared to ResNet50, suggesting that they are less influenced and remain strongly attached to local data representativeness. These two architectures are an ideal choice in a context where a compromise between minimizing communication costs and network stability, while maintaining acceptable performance, is a priority. Local learning without external influence (scenario where $\lambda = 0$) remains effective, but more unstable and slightly less efficient compared to scenarios using external information via the $\lambda$-weighted $Loss_{val}$ for local control. However, although the local integration of external information in the learning correction process is beneficial overall in terms of performance improvement, its effect varies depending on the architectures. It is therefore preferable to adjust $\lambda$ during the local learning correction process via the external $Loss_{val}$ depending on the model used.

\subsubsection{Impact of the number of shared models}
The aim of this study is to evaluate, on the basis of this new decentralized federated learning framework, how the number of models shared between clients influences the performance of each of the three architectures tested: ResNet50, VGG16 and ViT\_B16. To do this, we used a configuration of 6 clients with two model sharing scenarios, namely: a scenario of sharing one (1) best model and a scenario of sharing 5 best models between the clients participating in the process. Table~\ref{tab:SharedModels} shows the mean and standard deviation values of F1-score and accuracy for our three architectures as a function of $\lambda$, in a 5-best model sharing scenario. For this study, we used the same experimental parameters as in the previous study (i.e., a 6-client configuration, 50 cycles of communication between clients, a local epoch for each client and the same data partitioning policy). We will perform a comparative analysis between the results in Table~\ref{tab:lambdaImpact} (single best model sharing scenario) and those in Table~\ref{tab:SharedModels} (5 best models sharing scenario) to measure the impact of the number of shared models on model performance.

\begin{table*}[t]
\centering
\caption{Results from the study of the number of shared models, in 6-client configuration, five best shared models. }\label{tab:SharedModels}
\begin{tabular}{|c|cc|cc|cc|}
\hline
\multirow{2}{*}{\textbf{$\lambda$}} & \multicolumn{2}{c|}{\textbf{ResNet50}}                                 & \multicolumn{2}{c|}{\textbf{VGG16}}                                    & \multicolumn{2}{c|}{\textbf{ViT\_B16}}                                 \\ \cline{2-7} 
                            & \multicolumn{1}{c|}{\textbf{F1-Score}}       & \textbf{Accuracy}       & \multicolumn{1}{c|}{\textbf{F1-Score}}       & \textbf{F1-Score}       & \multicolumn{1}{c|}{\textbf{Accuracy}}       & \textbf{F1-Score}       \\ \hline
\textbf{0.00}               & \multicolumn{1}{c|}{70,37 ± 10,09}           & 75,78 ± 9,87            & \multicolumn{1}{c|}{99,23 ± 0,49}            & 99,23 ± 0,49            & \multicolumn{1}{c|}{97,52 ± 1,40}            & 97,55 ± 1,36            \\ \hline
\textbf{0.25}               & \multicolumn{1}{c|}{96,88 ± 4,75}            & 96,97 ± 4,61            & \multicolumn{1}{c|}{\textbf{99,33 ±   0,68}} & \textbf{99,34 ±   0,66} & \multicolumn{1}{c|}{97,74 ± 1,83}            & 97,79 ± 1,76            \\ \hline
\textbf{0.5}                & \multicolumn{1}{c|}{\textbf{97,77 ±   4,03}} & \textbf{97,81 ±   3,95} & \multicolumn{1}{c|}{98,57 ± 0,85}            & 98,59 ± 0,85            & \multicolumn{1}{c|}{97,64 ± 1,47}            & 97,70 ± 1,38            \\ \hline
\textbf{0.75}               & \multicolumn{1}{c|}{95,52 ± 7,86}            & 96,02 ± 7,05            & \multicolumn{1}{c|}{99,01 ± 0,80}            & 99,01 ± 0,81            & \multicolumn{1}{c|}{\textbf{98,23 ±   1,30}} & \textbf{98,23 ±   1,29} \\ \hline
\end{tabular}
\end{table*}

Performing a comparative analysis between the results in Table~\ref{tab:lambdaImpact} and those in Table~\ref{tab:SharedModels}, the ResNet50 model benefits enormously from sharing 5 best models by reducing variability, especially for $\lambda = 0.25$ and $\lambda = 0.50$, where the mean F1-score (case $\lambda = 0. 25$) rises from 91.14\% ± 11.37 for 1 best shared model to 96.88\% ± 4.75 for 5 best models; then from 93.21\% ± 8.92 (1 best shared model) to 97.77\% ± 4.03 (5 best shared models) in the case where $\lambda = 0.50$. This suggests that this architecture requires greater data variability in order to achieve more consistent performance. The VGG16 model achieves very good performance and is less affected than ResNet50 by the effect of the number of shared models. Its performance is very stable, with a slight peak at $\lambda = 0.25$, where the average F1-score rises from 98.91\% ± 1.35 (1 best model shared) to 99.33\% ± 0.68 (5 best models shared). While sharing 5 best models in the case of VGG16 slightly reduces variability, the difference in performance with the scenario of sharing a single best model remains insignificant. This suggests the robustness of this architecture to the number of shared models, and would therefore make it an optimal choice in a context where minimizing communication costs and network stability, while maintaining very good performance, is a priority. Like VGG16, ViT\_B16 achieves good performance and is also less affected by the number of shared models than ResNet50. Its performance peaks at $\lambda = 0.75$, where the average F1-score rises from 97.37\% ± 1.57 (1 best shared model) to 98.23\% ± 1.30 (5 best shared models). Although its performance is slightly inferior to that of VGG16, ViT\_B16 shows a modest performance improvement and better stability in the 5 shared models scenario, with a slight performance difference compared to the single shared model scenario. This suggests that this architecture would be a reasonable choice in a context where slightly lower performance than VGG16, but relatively stable with reduced communication costs, is a priority.

\begin{figure*}[t]
    \centering
    \includegraphics[width=\linewidth]{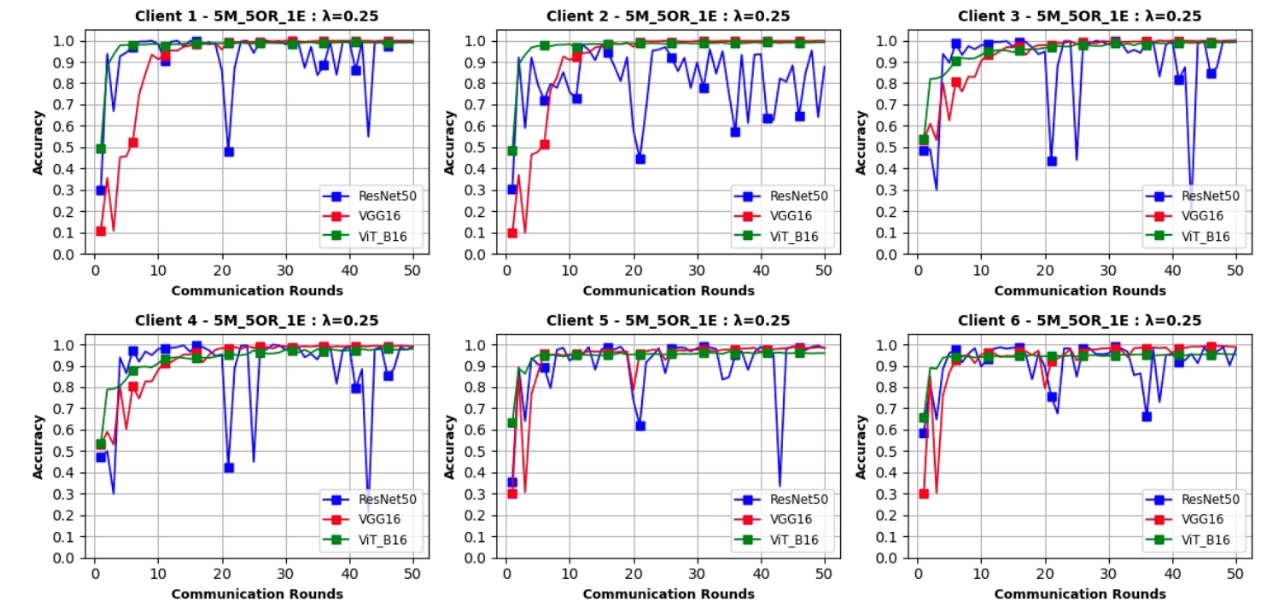}
    \caption{Impact of the number of shared models, in 6-client configuration, five best shared models, with $\lambda = 0.25$.}
    \label{fig:SharedModels6}
\end{figure*}

Figure~\ref{fig:SharedModels6} shows the evolution of the accuracy of the three models tested for $\lambda = 0.25$, according to each of the six clients after 50 rounds of communication, in a scenario of sharing five best models between clients. Compared with Figure~\ref{fig:SharedModel1}, for a scenario involving the sharing of a single best model, we can see a similar trend towards improved performance for our three models, with strong instability over the rounds for the ResNet50 model. VGG16 and ViT\_B16 show, in general, high performance stability, with ViT\_B16 showing a very rapid rise in performance and convergence after just 10 rounds, for the majority of customers. This suggests, in contrast to ResNet50, a good robustness of VGG16 and ViT\_B16 to data diversity via multiple model sharing, while being less influenced by external information with a small value of $\lambda$. In a context where minimizing communication costs and obtaining good stable performance, slightly lower than that of VGG16, is a priority, ViT\_B16 would be an optimal choice.

In summary, it can be seen that moving from a single-best model sharing scenario to a five-best model sharing scenario has a varied impact on the performance of our studied architectures. The scenario of sharing five best models improves stability and enables more homogeneous convergence of the local models of all the architectures studied. The ResNet50 model is the most affected by the number of shared models. It requires more than one best shared model to achieve more homogeneous performance, and would therefore not be an ideal choice in a scenario where minimizing communication costs is a priority. The VGG16 and ViT\_B16 models are the least impacted by the number of shared models, while maintaining very good performance and stability. They would therefore be an ideal choice in a context where a good compromise between minimizing communication costs and stable performance is a priority, with a preference for VGG16 due to its superior performance.

\subsubsection{Impact of the number of clients}
In this section, we study the impact of the number of clients participating in the decentralized federated learning process on the performance of each of the three architectures tested using this newly implemented learning framework, and as a function of each value of $\lambda$. To do this, we will compare the results obtained by moving from a six-client configuration to an 18-client configuration. To ensure equivalent comparison conditions, we consider, for both configurations, a scenario of sharing five best models between process participants, 50 cycles of communication between clients, one local epoch for each client and $Loss_{val}$ as the criterion for selecting and sharing the five best models. The results in Table~\ref{tab:ClientsNumber} represent the mean values and standard deviations of F1-score and accuracy obtained in the case of an 18-customer configuration. These results will be compared with those of the previous table (Table~\ref{tab:SharedModels}), which also correspond to mean values and standard deviations of F1-score and accuracy in the case of a six-client configuration.

\begin{table*}[t]
\centering
\caption{Results from the study of the impact of the number of clients. }\label{tab:ClientsNumber}
\begin{tabular}{|c|ll|ll|ll|}
\hline
\multirow{2}{*}{\textbf{$\lambda$}} & \multicolumn{2}{c|}{\textbf{ResNet50}}                                              & \multicolumn{2}{c|}{\textbf{VGG16}}                                                 & \multicolumn{2}{c|}{\textbf{ViT\_B16}}                                              \\ \cline{2-7} 
                            & \multicolumn{1}{c|}{\textbf{F1-Score}}     & \multicolumn{1}{c|}{\textbf{Accuracy}} & \multicolumn{1}{c|}{\textbf{F1-Score}}     & \multicolumn{1}{c|}{\textbf{Accuracy}} & \multicolumn{1}{c|}{\textbf{F1-Score}}     & \multicolumn{1}{c|}{\textbf{Accuracy}} \\ \hline
\textbf{0.00}               & \multicolumn{1}{l|}{77,26   ± 8,79}        & 76,91   ± 9,47                         & \multicolumn{1}{l|}{97,94   ± 1,52}        & 98,01   ± 1,43                         & \multicolumn{1}{l|}{97,94 ± 1,52}          & 98,01 ± 1,43                           \\ \hline
\textbf{0.25}               & \multicolumn{1}{l|}{98,71   ± 2,18}        & 98,79   ± 1,98                         & \multicolumn{1}{l|}{\textbf{98,47 ± 0,81}} & \textbf{98,48 ± 0,81}                  & \multicolumn{1}{l|}{\textbf{98,47 ± 0,81}} & \textbf{98,48 ± 0,81}                  \\ \hline
\textbf{0.50}               & \multicolumn{1}{l|}{\textbf{99,13 ± 0,86}} & \textbf{99,14 ± 0,86}                  & \multicolumn{1}{l|}{97,91   ± 1,42}        & 97,90   ± 1,45                         & \multicolumn{1}{l|}{97,91 ± 1,42}          & 97,90 ± 1,45                           \\ \hline
\textbf{0.75}               & \multicolumn{1}{l|}{98,76   ± 1,09}        & 98,77   ± 1,05                         & \multicolumn{1}{l|}{97,61   ± 1,03}        & 97,62   ± 1,03                         & \multicolumn{1}{l|}{97,61 ± 1,03}          & 97,62 ± 1,03                           \\ \hline
\end{tabular}
\end{table*}

According to the results obtained, by varying the number of clients from six (results in Table~\ref{tab:SharedModels}) to 18 (results in Table~\ref{tab:ClientsNumber}), the impact of this variation can be observed in different ways on the performance of our three models tested. 

The ResNet50 model benefits greatly from the increase in the number of clients, more so than VGG16 and ViT\_B16, with a marked improvement in performance and clear stability, whatever the value of $\lambda$. It reaches a peak in average performance in the case where $\lambda = 0.50$, with average accuracy rising from 97.81\% ± 3.95 to 99.14\% ± 0.86, respectively in the case of a six-client and 18-client configuration. This represents a remarkable reduction in variability, suggesting that ResNet50 takes advantage of data diversity with a large number of clients to improve its performance.

In contrast to ResNet50, VGG16 achieves good average performance in both configurations, but suffers a slight decrease in performance with 18 clients, with maximum average accuracy falling from 99.34\% ± 0.66 to 98.48\% ± 0.81, respectively in the case of a six-client and 18-client configuration, for $\lambda = 0.25$. This suggests that, despite its good stability, increasing the number of customers in the case of VGG16 leads to an information dispersion effect that does not benefit this model as much.

Increasing the number of clients in the case of ViT\_B16 slightly improves its performance. In the 18-client configuration, it reaches the peak of its average performance in the case where $\lambda = 0.25$, with average accuracy rising from 97.79\% ± 1.76 to 98.48\% ± 0.81, respectively in the six-client and 18-client configurations, suggesting that this model benefits from data diversity with more clients, unlike VGG16. Although its performance is stable and similar to that of VGG16, it does not improve as much as that of ResNet50 with more clients.

In summary, increasing the number of clients from six to 18 has a different impact on the performance of the models tested. The ResNet50 model benefits most from this increase, with a very noticeable improvement in performance. This suggests that, in a context where high performance is a priority, regardless of communication costs, ResNet50 would be an ideal choice. VGG16 and ViT\_B16 achieve similar, stable average performance, but with opposite effects. VGG16 would be ideal in a configuration with fewer clients; however, ViT\_B16 remains globally robust, whatever the configuration.

\subsubsection{Impact of using $Loss_{val}$ or $Loss_{train}$}
In this section, we have studied how the choice of sharing $Loss_{val}$ or $Loss_{train}$ influences the local learning correction process, and how it, impacts the performance of our three tested architectures. Indeed, according to equation (5) of phase 5 of our proposed approach, the adjustment of the local loss function is carried out by integrating, in the calculation, the loss $Loss_{val}$ of the best model received, weighted by the parameter $\lambda$. Thus, we considered the 18-customer configuration and used, in a first scenario, $Loss_{val}$, and in a second scenario, $Loss_{train}$ of the best model in the calculation of the new local loss function. The results in Table~\ref{tab:ValTrainLoss} summarize the mean values and standard deviations of F1-score and accuracy obtained in the scenario using $Loss_{train}$. These results will be compared with those of the previous table (Table~\ref{tab:ClientsNumber}), which also correspond to mean values and standard deviations of F1-score and accuracy, this time in the $Loss_{val}$ scenario. To ensure fair comparison conditions, we have considered, in both scenarios, the case of sharing five best models between process participants, 50 cycles of communication between customers, one local epoch for each customer and $Loss_{val}$ as the criterion for selecting and sharing the five best models.

\begin{table*}[t]
\centering
\caption{Results from the study of the impact of using $Loss_{train}$, in 18-client configuration, five best shared models. }\label{tab:ValTrainLoss}
\begin{tabular}{|c|cc|cc|cc|}
\hline
\multirow{2}{*}{\textbf{$\lambda$}} & \multicolumn{2}{c|}{\textbf{ResNet50}}                                 & \multicolumn{2}{c|}{\textbf{VGG16}}                                    & \multicolumn{2}{c|}{\textbf{ViT\_B16}}                                 \\ \cline{2-7} 
                            & \multicolumn{1}{c|}{\textbf{F1-Score}}       & \textbf{Accuracy}       & \multicolumn{1}{c|}{\textbf{F1-Score}}       & \textbf{Accuracy}       & \multicolumn{1}{c|}{\textbf{F1-Score}}       & \textbf{Accuracy}       \\ \hline
\textbf{0.00}               & \multicolumn{1}{c|}{97,96 ± 4,47}            & 97,99 ± 4,35            & \multicolumn{1}{c|}{97,04 ± 2,02}            & 97,14 ± 1,84            & \multicolumn{1}{c|}{97,04 ± 2,02}            & 97,14 ± 1,84            \\ \hline
\textbf{0.25}               & \multicolumn{1}{c|}{\textbf{98,81 ±   1,09}} & \textbf{98,81 ±   1,07} & \multicolumn{1}{c|}{\textbf{98,46 ±   0,67}} & \textbf{98,46 ±   0,68} & \multicolumn{1}{c|}{\textbf{98,46 ±   0,67}} & \textbf{98,46 ±   0,68} \\ \hline
\textbf{0.50}               & \multicolumn{1}{c|}{65,12 ± 8,21}            & 66,30 ± 7,37            & \multicolumn{1}{c|}{98,18 ± 0,77}            & 98,14 ± 0,90            & \multicolumn{1}{c|}{98,18 ± 0,77}            & 98,14 ± 0,90            \\ \hline
\textbf{0.75}               & \multicolumn{1}{c|}{73,43 ± 8,00}            & 75,16 ± 8,26            & \multicolumn{1}{c|}{96,16 ± 2,67}            & 96,15 ± 2,72            & \multicolumn{1}{c|}{96,16 ± 2,67}            & 96,15 ± 2,72            \\ \hline
\end{tabular}
\end{table*}

Comparing the results in Table~\ref{tab:ClientsNumber} ($Loss_{val}$ scenario) with those in Table~\ref{tab:ValTrainLoss} ($Loss_{train}$ scenario), we can see that, in the $Loss_{val}$ scenario, all three models tested achieve better, overall more stable performance than in the $Loss_{train}$ case, where we observe a negative impact, particularly in the case where $\lambda = 0.75$, regardless of architecture.

The ResNet50 model achieves better mean high performance with $Loss_{val}$ for $\lambda \geq 0.50$, with a peak of 99.14\% ± 0.86 accuracy and very good stability in the case where $\lambda = 0.50$. However, under the same conditions ($\lambda = 0.50$) and with the scenario of using $Loss_{train}$, ResNet50 suffers a sharp degradation in performance, reaching a lower average accuracy of 66.30\% ± 7.37, suggesting that using $Loss_{val}$ seems to be a better choice for this architecture, as it achieves better performance while ensuring better generalization.

The VGG16 model achieves overall stable performance, with minimal differences between the $Loss_{val}$ and $Loss_{train}$ scenarios, whatever the value of $\lambda$. It reached a performance peak of 98.48\% ± 0.81 average accuracy versus 98.46\% ± 0.67 average accuracy in the $Loss_{val}$ and $Loss_{train}$ scenarios respectively, suggesting the robustness of this architecture in the face of data diversity. However, the use of $Loss_{val}$ remains a better option, as it delivers better performance and more balanced convergence, while helping to obtain a more representative overall model.

The ViT\_B16 model achieves very similar results to VGG16, with a slight drop in performance for $\lambda = 0.75$. It reaches a peak performance of 98.47\% ± 0.81 of mean F1-score versus 98.46\% ± 0.67 of mean F1-score, respectively in the scenario using $Loss_{val}$ and that using $Loss_{train}$, suggesting, as for VGG16, good robustness. However, for more robust convergence and better performance stability, $Loss_{val}$ remains preferable.

\begin{figure*}[t]
    \centering
    \includegraphics[width=\linewidth]{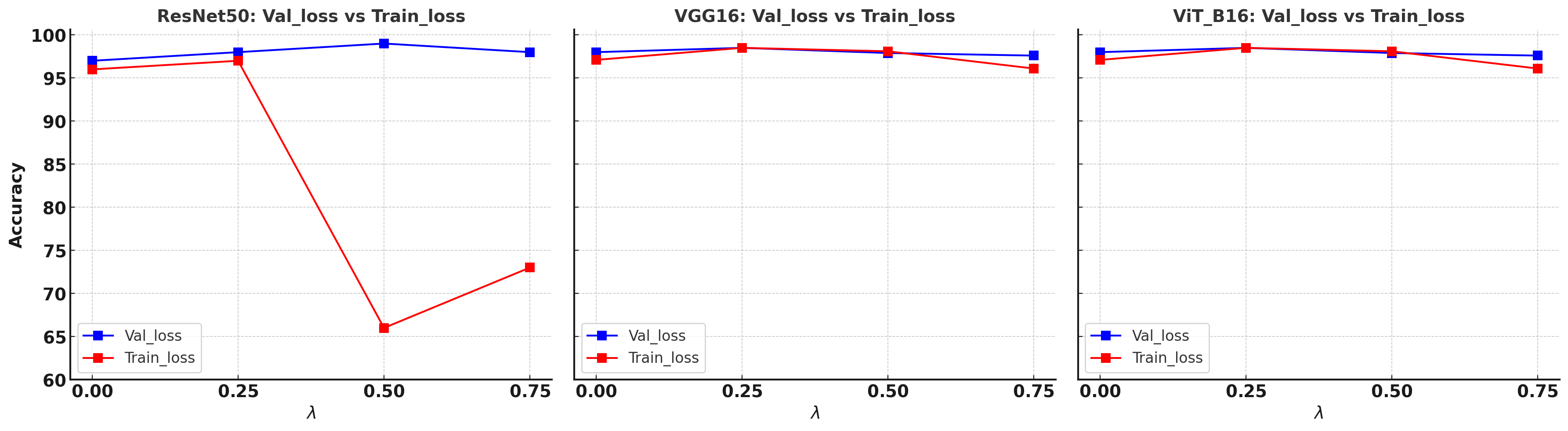}
    \caption{Impact of using $Loss_{val}$ vs $Loss_{train}$ in a 18-client configuration, five best shared models.}
    \label{fig:ValTrainLoss}
\end{figure*}

Figure~\ref{fig:ValTrainLoss} shows the performance trend (average accuracy) of each of our three tested models as a function of $\lambda$ for the two scenarios studied (the scenario using $Loss_{val}$ and the one using $Loss_{train}$) in the local learning correction process via the calculation of the new loss function. We can see that the use of $Loss_{train}$ leads to a sharp degradation in performance in the case of ResNet50, for $\lambda \geq 0.50$. On the other hand, in the case of VGG16 and ViT\_B16, the use of $Loss_{train}$ generates a slight drop in performance for $\lambda = 0.75$, with minimal difference from the scenario using $Loss_{val}$, confirming the robustness of these two models. Whatever the model, we find that using $Loss_{val}$ generally ensures stable performance, good generalization and guarantees more balanced convergence, making its use in this context a better choice.

In summary, from the comparative analysis of performance results between the $Loss_{val}$ and $Loss_{train}$ scenarios for the five best models shared in the local learning correction process via the calculation of the new loss function, we can see that the $Loss_{val}$ scenario ensures more stable and higher overall performance for our three studied architectures. The ResNet50 model benefits most from $Loss_{val}$, while with $Loss_{train}$, its performance drops drastically. The VGG16 and ViT\_B16 models show similar robustness in both scenarios, with a slight superiority of $Loss_{val}$ for better stability and more balanced convergence. Overall, the scenario using $Loss_{val}$ appears to be a more relevant choice, as it favors better generalization and a more representative federated model. Moreover, as $Loss_{train}$ is the measure of model error on training data, and in the context of federated learning, clients generally have heterogeneous data, suggesting that the use of $Loss_{train}$ in local learning correction may be biased, as it depends solely on local training data, already known to the client model. However, incorporating $Loss_{val}$ into local learning correction ensures that local models improve by taking into account their ability to predict well on unseen data, thus reducing the risk of overlearning. This also enables each client to adjust its learning based on global rather than local performance, improving cooperation and consistency of model updates.

The results obtained through our decentralized federated learning framework are promising, and the proposed approach constitutes a dynamic and robust solution, while preserving the confidentiality of clients' private data. The elimination of dependency on a central server for aggregation of local models makes this framework particularly suited to the agricultural domain, especially in rural areas or resource-constrained environments. Furthermore, the $Loss_{val}$-based strategy of model sharing and local learning correction guarantees, on the one hand, the propagation of high-quality models and, on the other, the correction of local learning through controlled regulation of the influence of external models via the $\lambda$ parameter. This accelerates global model convergence, while reducing communication costs and improving model accuracy.

Comparing our results with those presented in the literature, notably in \cite{G2019}, \cite{Chen2022} for pre-trained CNNs using the classical or traditional machine learning approach, and in \cite{MambaKabala2023} for pre-trained CNNs and ViTs using the standard or centralized federated learning approach, and as summarized in Table~\ref{tab:Comparison}, we find that our proposed framework achieves superior performance to that reported in these works, using the same datasets from the PlantVillage platform. Furthermore, with fewer clients or participants in the learning process, and keeping the same experimental parameters, our proposed approach achieves superior performance for VGG16 and ViT\_B16, and almost similar for ResNet50 as those obtained in \cite{MambaKabala2023}. This demonstrates the robustness of our approach, in addition to the many advantages it offers, both in terms of security and confidentiality of participants' data, and in terms of high performance achieved at low communication cost.

\begin{table*}[t]
\centering
\caption{Comparison of results between our proposed approach and the literature. }\label{tab:Comparison}
\begin{tabular}{|l|cc|cc|cc|cc|}
\hline
\multirow{2}{*}{} & \multicolumn{2}{c|}{\textbf{Our app. (client=6)}}       & \multicolumn{2}{c|}{\textbf{{\cite{MambaKabala2023}} (client=7)}}           & \multicolumn{2}{c|}{\textbf{{\cite{Chen2022}} }}                      & \multicolumn{2}{c|}{\textbf{{\cite{G2019}} }}                      \\ \cline{2-9} 
                  & \multicolumn{1}{c|}{\textbf{F1-score}} & \textbf{Acc.} & \multicolumn{1}{c|}{\textbf{F1-score}} & \textbf{Acc.} & \multicolumn{1}{c|}{\textbf{F1-score}} & \textbf{Acc.} & \multicolumn{1}{c|}{\textbf{F1-score}} & \textbf{Acc.} \\ \hline
\textbf{ResNet50} & \multicolumn{1}{c|}{99,00}             & 99,02             & \multicolumn{1}{c|}{\textbf{99,52}}    & \textbf{99,52}    & \multicolumn{1}{c|}{97,90}             & 98,60             & \multicolumn{1}{c|}{×}                 & 92,56             \\ \hline
\textbf{VGG16}    & \multicolumn{1}{c|}{\textbf{98,91}}    & \textbf{98,76}    & \multicolumn{1}{c|}{96,96}             & 96,94             & \multicolumn{1}{c|}{96,80}             & 97,50             & \multicolumn{1}{c|}{×}                 & 92,87             \\ \hline
\textbf{ViT\_B16} & \multicolumn{1}{c|}{\textbf{97,71}}    & \textbf{97,55}    & \multicolumn{1}{c|}{94,83}             & 94,80             & \multicolumn{1}{c|}{×}                 & ×                 & \multicolumn{1}{c|}{×}                 & ×                 \\ \hline
\end{tabular}
\end{table*}
Our framework can be implemented by farmers or agricultural cooperatives to train global models from different datasets from various regions, with the aim of improving model robustness and accuracy in plant disease detection, diagnosis and classification tasks. Although our approach has many significant advantages, some challenges remain. Connectivity issues can lead to communication difficulties between clients, which can impact the effectiveness of the learning framework. In addition, the performance of the approach may also be affected in the presence of non-IDI (Independent and Identitically Distributed) data.

\section{Conclusion}\label{sec5}
In this work, we have proposed a novel decentralized federated learning (DFL) framework for crop disease classification that addresses key challenges related to confidentiality, communication efficiency and robustness in distributed learning environments. By relying on a peer-to-peer architecture and eliminating the need for a central server, our approach is particularly suited to agricultural applications where data sensitivity and connectivity constraints are common, especially in rural or resource-limited areas. The originality of our framework lies in the use of validation loss ($Loss_{val}$) as a guiding measure for model sharing and local learning correction. By introducing an adaptive loss function regulated by the parameter $\lambda$, we enable each client to adjust its local updates based not only on its own learning data, but also on the performance of the best models received from its peers. This provides a better balance between local data representativeness and global model generalization. We conducted extensive experiments using three pre-trained deep learning architectures : ResNet50, VGG16 and ViT\_B16 on the PlantVillage dataset. The results show that all three architectures benefit from external correction via $Loss_{val}$, but to varying degrees. ResNet50, while achieving high accuracy, seems highly sensitive to external influence and therefore less reliable in maintaining local data consistency. In contrast, VGG16 and ViT\_B16 show a more balanced behavior, combining robustness and high local representativeness, making them ideal candidates in scenarios prioritizing communication efficiency and model stability. Overall, our approach outperforms centralized FL and traditional CNN-based machine learning methods in the literature, even with fewer clients and under limited communication conditions. These results underline the viability of our DFL framework as a high-performance, privacy-friendly solution for collaborative crop disease classification in decentralized agricultural networks. In future work we will try to address the challenges of non-ID data distributions, scalability and real-world deployment testing of our solution.

\section*{Acknowledgements}
This work is funded as part of the MERIAVINO project. MERIAVINO is part of the ERA-NET Cofund ICT-AGRI-FOOD, with funding provided by national sources [ANR France, UEFISCDI Romania, GSRI Greece] and co-funding by the European Union’s Horizon 2020 research and innovation program, Grant Agreement number 862665.  

\section*{Authors contributions}
D.M.K. contributed to data collection, development of the methodological architecture, implementation of the models, execution of the various experiments, interpretation of the results and drafting of the manuscript. - A.H. coordinated the research work and conceptualized the method. - All authors discussed and validated the working methodology, the experiments carried out, the interpretation of results, and the definition of a plan for writing the manuscript. - A.H., L.B. and R.C. contributed to the reading and correction of the manuscript, both in terms of content and form. - All authors read and approved the final version of the manuscript.

\bibliographystyle{IEEEtran} 
\bibliography{Ref}
\end{document}